%% file: main.tex
\documentclass[conference]{IEEEtran}
\IEEEoverridecommandlockouts
\input{core.constants.tex}
\input{core.config.tex}
\usepackage{hyperref}
\input{core.include.tex}
\input{core.thm.env.tex}

\usepackage{subfig}
\usepackage{txfonts} 
\usepackage{lipsum}
\usepackage{adjustbox}
\usepackage{cite}
\usepackage{amsmath,amssymb,amsfonts}
\usepackage{algorithmic}
\usepackage{graphicx}
\usepackage{textcomp}
\usepackage{xcolor}
\usepackage{tikz}
\usepackage{subfig}
\usetikzlibrary{arrows.meta}
\usepackage{ulem}
\def\BibTeX{{\rm B\kern-.05em{\sc i\kern-.025em b}\kern-.08em
    T\kern-.1667em\lower.7ex\hbox{E}\kern-.125emX}}
\begin{document}

\title{\articletitle
}

\author{%
\IEEEauthorblockN{%
Zihang~Su\IEEEauthorrefmark{1}\href{https://orcid.org/0000-0003-1039-7462}{\protect\includegraphics[scale=0.05]{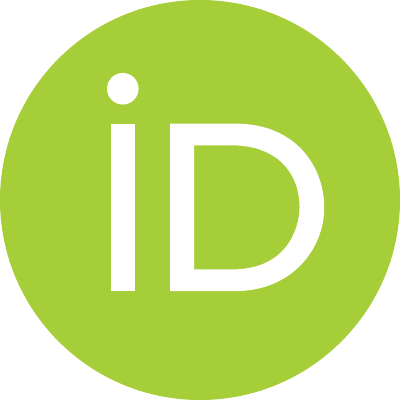}},
Tianshi~Yu\IEEEauthorrefmark{2}\href{https://orcid.org/0000-0002-3345-3028}{\protect\includegraphics[scale=0.05]{fig/orcid}},
Nir~Lipovetzky\IEEEauthorrefmark{1}\href{https://orcid.org/0000-0002-8667-3681}{\protect\includegraphics[scale=0.05]{fig/orcid}},
Alireza~Mohammadi\IEEEauthorrefmark{2}\href{https://orcid.org/0000-0002-5561-1322}{\protect\includegraphics[scale=0.05]{fig/orcid}}, 
Denny~Oetomo\IEEEauthorrefmark{2}\href{https://orcid.org/0000-0002-2680-6489}{\protect\includegraphics[scale=0.05]{fig/orcid}},\\
Artem~Polyvyanyy\IEEEauthorrefmark{1}\href{https://orcid.org/0000-0002-7672-1643}{\protect\includegraphics[scale=0.05]{fig/orcid}},
Sebastian~Sardi{\~{n}}a\IEEEauthorrefmark{3}\href{https://orcid.org/0000-0003-2962-0118}{\protect\includegraphics[scale=0.05]{fig/orcid}},
Ying~Tan\IEEEauthorrefmark{2}\href{https://orcid.org/0000-0001-8495-0246}{\protect\includegraphics[scale=0.05]{fig/orcid}},
Nick~van~Beest\IEEEauthorrefmark{4}\href{https://orcid.org/0000-0003-3199-1604}{\protect\includegraphics[scale=0.05]{fig/orcid}}
}%
\IEEEauthorblockA{%
\IEEEauthorrefmark{1}School of Computing and Information Systems, The University of Melbourne, Australia\\
\href{mailto:zihangs@student.unimelb.edu.au}{zihangs@student.unimelb.edu.au}, \href{mailto:nir.lipovetzky@unimelb.edu.au;artem.polyvyanyy@unimelb.edu.au}{\{nir.lipovetzky,artem.polyvyanyy\}@unimelb.edu.au}%
}
\IEEEauthorblockA{\IEEEauthorrefmark{2}Department of Mechanical Engineering, The University of Melbourne, Australia\\
\href{mailto:tianshiy@student.unimelb.edu.au}{tianshiy@student.unimelb.edu.au}, \href{mailto:alireza.mohammadi@unimelb.edu.au;doetomo@unimelb.edu.au;yingt@unimelb.edu.au}{\{alireza.mohammadi,doetomo,yingt\}@unimelb.edu.au}\\
}
\IEEEauthorblockA{\IEEEauthorrefmark{3}School of Computing Technologies, RMIT University, Australia\\
\href{mailto:sebastian.sardina@rmit.edu.au}{sebastian.sardina@rmit.edu.au}
}
\IEEEauthorblockA{\IEEEauthorrefmark{4}Software Systems Group, Data61 $\mid$ CSIRO, Australia\\
\href{mailto:nick.vanbeest@data61.csiro.au}{nick.vanbeest@data61.csiro.au}
}
}

\maketitle
\begin{abstract}
\input{tex/abstract}
\end{abstract}



\input{tex/introduction.tex}

\input{tex/Project_Description.tex}

\input{tex/related_work.tex}
\input{tex/approach.tex}
\input{tex/evaluation.tex}

\input{tex/discussion.tex}
\input{tex/conclusion.tex}

\bibliography{bibliography}
\end{document}

%% file: core.config.tex
\newcommand{\articletitle}         {Data-Driven Goal Recognition in Transhumeral Prostheses Using Process Mining Techniques}

\newif\ifshowtodos
\showtodostrue		



%% file: core.include.tex
\usepackage{booktabs}
\usepackage{multirow}
\usepackage{nicefrac}
\usepackage{ifthen}
\usepackage[usenames,dvipsnames]{xcolor}
\usepackage[algoruled,vlined,linesnumbered]{algorithm2e}
\usepackage{amsmath}
\usepackage{amsfonts}
\usepackage{mdframed}
\usepackage{paralist}
\setdefaultleftmargin{1em}{1em}{1em}{1em}{1em}{1em}
\usepackage{datetime}
\newdateformat{monthyeardate}{\monthname[\THEMONTH] \THEYEAR}
\usepackage[algoruled]{algorithm2e}
\ifshowtodos
\usepackage[textsize=scriptsize,colorinlistoftodos]{todonotes}
\else
\usepackage[textsize=scriptsize,colorinlistoftodos,disable]{todonotes} 
\fi
\let\todonote\todo
\renewcommand{\todo}[2]{\todonote[inline,color=red!20]{TODO (#1): #2}}

\usepackage{tikz,pgf,xcolor}
\usetikzlibrary{arrows,automata,trees,plotmarks,shadows,shapes,decorations.pathmorphing,backgrounds,positioning,fit,petri}
\usepackage{pgfplots}
\usetikzlibrary{pgfplots.groupplots}
\usepackage{upgreek}
\usepackage{mathtools} 
\usepackage[capitalise,nameinlink]{cleveref}
\usepackage{wrapfig}

\definecolor{mybluecolor}{RGB}{50,106,218}
\definecolor{myredcolor}{RGB}{176,53,53}
\definecolor{mygreencolor}{RGB}{93,172,0}
\definecolor{myyellowcolor}{RGB}{255,163,34}
\definecolor{mypurplecolor}{RGB}{86,35,132}
\definecolor{mytealcolor}{RGB}{30,161,165}

\newcommand{\splitatcommas}[1]{%
  \begingroup
  \ifnum\mathcode`,="8000
  \else
    \begingroup\lccode`~=`, \lowercase{\endgroup
      \edef~{\mathchar\the\mathcode`, \penalty0 \noexpand\hspace{-1pt plus 3em}}%
    }\mathcode`,="8000
  \fi
  #1%
  \endgroup
}

\newcommand\fixithelp[2]{%
  \wd0=\dimexpr\wd0-\linewidth\relax%
  \ifdim\wd0>0pt\relax%
    \fixithelp{#1}{#2}%
  \else%
    \wd0=\dimexpr\wd0+\linewidth\relax
    \ifdim\wd0>.9\linewidth\relax%
      {\parfillskip0pt\relax#2\par}%
    \else%
      \ifdim\wd0>.8\linewidth\relax%
        {\parfillskip0pt\relax#2\hspace{.2\linewidth}\par}%
      \else%
        \ifdim\wd0<#1\linewidth\relax%
          {\parfillskip0pt\relax#2\par}%
        \else%
          \ifdim\wd0<.2\linewidth\relax%
            {\parfillskip0pt\relax#2\hspace{.8\linewidth}\mbox{}\par}%
          \else%
            #2%
          \fi
        \fi
      \fi
    \fi
  \fi%
}

%% file: core.thm.env.tex
\newtheorem{mytheorem}		{Theorem}
\newtheorem{mydefinition}	{Definition}
\newtheorem{mylemma}			{Lemma}
\newtheorem{myproposition}{Proposition}
\newtheorem{mycorollary}	{Corollary}
\newtheorem{myexample}		{Example}
\newtheorem{myconjecture}	{Conjecture}

\newtheorem{myinvariant}	{Invariant}

\numberwithin{mytheorem}		{section}
\numberwithin{mydefinition}	{section}
\numberwithin{mylemma}			{section}
\numberwithin{myproposition}{section}
\numberwithin{mycorollary}	{section}
\numberwithin{myexample}		{section}
\numberwithin{myconjecture}	{section}
\numberwithin{myremark}			{section}
\numberwithin{myinvariant}	{section}

\addtocounter{mytheorem}{0}


%% file: tex/abstract.tex
A transhumeral prosthesis restores missing anatomical segments below the shoulder, including the hand. Active prostheses utilize real-valued, continuous sensor data to recognize patient target poses, or goals, and proactively move the artificial limb. Previous studies have examined how well the data collected in stationary poses, without considering the time steps, can help discriminate the goals. In this case study paper, we focus on using time series data from surface electromyography electrodes and kinematic sensors to sequentially recognize patients' goals. Our approach involves transforming the data into discrete events and training an existing process mining-based goal recognition system. Results from data collected in a virtual reality setting with ten subjects demonstrate the effectiveness of our proposed goal recognition approach, which achieves significantly better precision and recall than the state-of-the-art machine learning techniques and is less confident when wrong, which is beneficial when approximating smoother movements of prostheses.

%% file: tex/introduction.tex
\section{Introduction}
\label{sec:introduction}

Given a collection of candidate goals and observations of actions performed by an agent in an environment, a solution to the goal recognition (GR) problem suggests the true goal the agent strives to achieve in the environment~\cite{mirsky2021introduction}.
Process mining (PM) techniques were recently used to implement a system for solving the GR problem~\cite{DBLP:conf/atal/PolyvyanyySLS20,AIJ_Su}.
We refer to this system as the PM-based GR system.
Assuming the actions the agent uses to achieve the various goals do not vary significantly over time, the PM-based GR system uses \emph{process discovery} techniques~\cite{DBLP:conf/icpm/LeemansPW19} to construct process models, or behavior models, from the action sequences the agent performed in the past to achieve the goals. 
Each model is constructed from the historical observations of how the agent reached a specific goal and represents the exemplary behavior for achieving the goal.
Subsequently, the PM-based GR system uses \emph{conformance checking} techniques~\cite{DBLP:journals/widm/AalstAD12} to align a newly observed action sequence (a trace) with each discovered behavior model. 
Finally, the obtained conformance information regarding the commonalities and discrepancies between the trace and all the models is translated into a probability distribution over the candidate goals. 
This distribution describes the probabilities of the agent pursuing the goals.

Goal recognition techniques are used in many real-world scenarios, such as autonomous driving~\cite{DBLP:conf/iros/BrewittGGA21,DBLP:conf/icra/AlbrechtBWGEDR21}, robotics~\cite{DBLP:journals/cp/Demiris07,DBLP:conf/iros/0002SKC0K20}, and human-machine interaction~\cite{DBLP:journals/aim/RichSL01}.
This paper investigates the feasibility and potential benefits of using the PM-based GR system in active transhumeral prostheses designed to assist individuals with disabilities~\cite{Prost:Legrand2022,Prost:Yu2023,DBLP:conf/smc/YuG00CO21}.
A transhumeral prosthesis replaces the function of missing anatomical segments below the shoulder of a patient, including the hand.
An active prosthesis is equipped with sensors and motors that aim to recognize the intents of the patient and support them, for example, by automatically directing the artificial limb towards the target prosthetic pose, namely the goal.

This study uses the dataset collected for the development of active transhumeral prostheses. The data collection experiment involved ten non-disabled subjects in a virtual reality (VR) environment, aiming to emulate the behavior of patients using transhumeral prostheses~\cite{Prost:Yu2023}.
Each subject was requested to accomplish forward-reaching tasks that involved achieving three elbow poses, namely the goals. They were required to extend their sound upper limb forward in a series of 30 iterations for each goal. The reaching targets are placed along the parasagittal plane.
A small sphere in the VR environment communicated a reaching target to the subject, and the subject had to reach it with their hand.
The subjects were requested to stay still for approximately one second after reaching the goal.
During each iteration towards a goal, 47 features were extracted from the measurements of kinematic and surface electromyography (sEMG) sensors attached to the subject at regular intervals.
These sequences of features constitute time series of continuous, real-valued data 
that characterize the behavior of the subject in achieving the goal.

This work is a natural extension of our previous research, which demonstrated that the measurements during the one-second holding period contain features suitable for discriminating between different fixed poses of the subjects, without considering the sequence data~\cite{Prost:Yu2023}. However, intuitively, considering the time stamps in the measurements is expected to improve performance. In this study, we investigate the potential of utilizing sequences of these measurements to infer the subjects' goal poses as they strive to achieve them. By predicting the patient's goals, our aim is to inform the design of active prostheses that facilitate cooperation between the patient and the prosthesis, ultimately enhancing the overall user experience. To accomplish goal inference, we leverage our PM-based GR system. Additionally, since the GR system operates on discrete event data, we propose a two-stage approach for transforming the multi-dimensional, real-valued, continuous measurements into discrete events.

We compared the performance of the PM-based GR system with two state-of-the-art GR methods in the area of active prostheses: linear discriminant analysis (LDA)~\cite{DBLP:conf/smc/YuG00CO21} and long short-term memory (LSTM) neural networks~\cite{huang2021development}.
The results confirm that the PM-based GR system outperforms the LSTM-based and LDA-based machine learning methods.
The artifacts constructed by the PM-based system (discovered models and alignments between new observations and the models) can be used to explain the inferences.
Furthermore, the PM-based GR tends to provide less confident results when making mistakes, which is beneficial when approximating smoother movements of prostheses.
The practical solution to the GR problem in the field of transhumeral prostheses can benefit from further developments in the GR techniques and their ensembling.
Specifically, this paper contributes:
\smallskip
\begin{compactitem}
\item 
An extension of the data-driven approach for goal recognition grounded in process mining techniques~\cite{DBLP:conf/atal/PolyvyanyySLS20,AIJ_Su} to repeated multi-dimensional, real-valued, continuous measurements that characterize the observed behavior of interest;
\item 
The results of an evaluation based on a publicly available implementation\footnote{\href{https://doi.org/10.26188/24131493}{https://doi.org/10.26188/24131493}} of the data-driven GR systems discussed in this paper, including the one grounded in process mining techniques, that compares their performance over a publicly available dataset\footnote{\href{https://doi.org/10.26188/23294693}{https://doi.org/10.26188/23294693}} in the domain of transhumeral prostheses. These results confirm that the process mining approach achieves significantly better precision and recall in recognizing the target patients’ poses and is less confident when wrong than the state-of-the-art machine learning techniques.
\end{compactitem}
\smallskip

\noindent
The next section introduces our transhumeral prosthetic study.
Next, \cref{sec:related:work} discusses related work.
\Cref{sec:approach} presents the extended version of the PM-based GR system, which is evaluated against state-of-the art machine learning approaches in
\Cref{sec:evaluation}.
Finally, \cref{sec:discussion} discusses the limitations of the current techniques and future work aiming to address these limitations, while \cref{sec:conclusion} concludes this paper.

%% file: tex/Project_Description.tex
\section{Study Description}
\label{sec:project:description}

\subsection{Background}
    The main objective of our study is to develop active transhumeral prostheses that replace missing limb segments below the shoulder, restoring upper limb function for achieving specific prosthetic poses. Active prostheses are robotic devices with actuators, like electric motors, driving prosthetic joints for tasks such as reaching. In this work, these joint movements are controlled by synthesizing data from above-elbow sEMG and joint movement sensors on the residual limb (upper arm) and body~\cite{Prost:Legrand2022,Prost:Yu2023}. Given that the user-intended prosthetic joint pose is unknown to the prosthetic device, the precise identification of this goal becomes crucial. Failure to achieve this can lead to inefficient task execution, user dissatisfaction, and even potential abandonment of the device~\cite{Prost:Hargrove2010}. However, it is challenging to develop goal recognition to accurately detect the intended goal in prosthetic settings due to significant human variance in kinematic and muscle activity signals~\cite{Prost:Shehata2021}.

    The goal recognition algorithm is required to be capable of accurately identifying the intended goals (target prosthetic poses) based solely on the signals accessible above the elbow. To this end, it is a common practice to first explore datasets from non-disabled human subjects~\cite{Prost:Hargrove2010}. These datasets capture the above-elbow sEMG and joint movement data for every sampling time instance while the subjects perform the task of forward-reaching. Each of these instances is labeled with its corresponding goal. The studied dataset herein was collected from ten non-disabled subjects performing forward-reaching tasks by extending their intact upper limb forward in a head-mounted display (HMD) VR environment. The data collection focused on transhumeral prostheses with a single prosthetic elbow joint. The aim was to attain distinct elbow poses that corresponded to specific spatial locations for forward-reaching. 

\begin{figure*}[t]
\vspace{-2mm}
\centering
\subfloat[]{%
\includegraphics[width=0.27\textwidth, trim=0 0 0 3, clip]{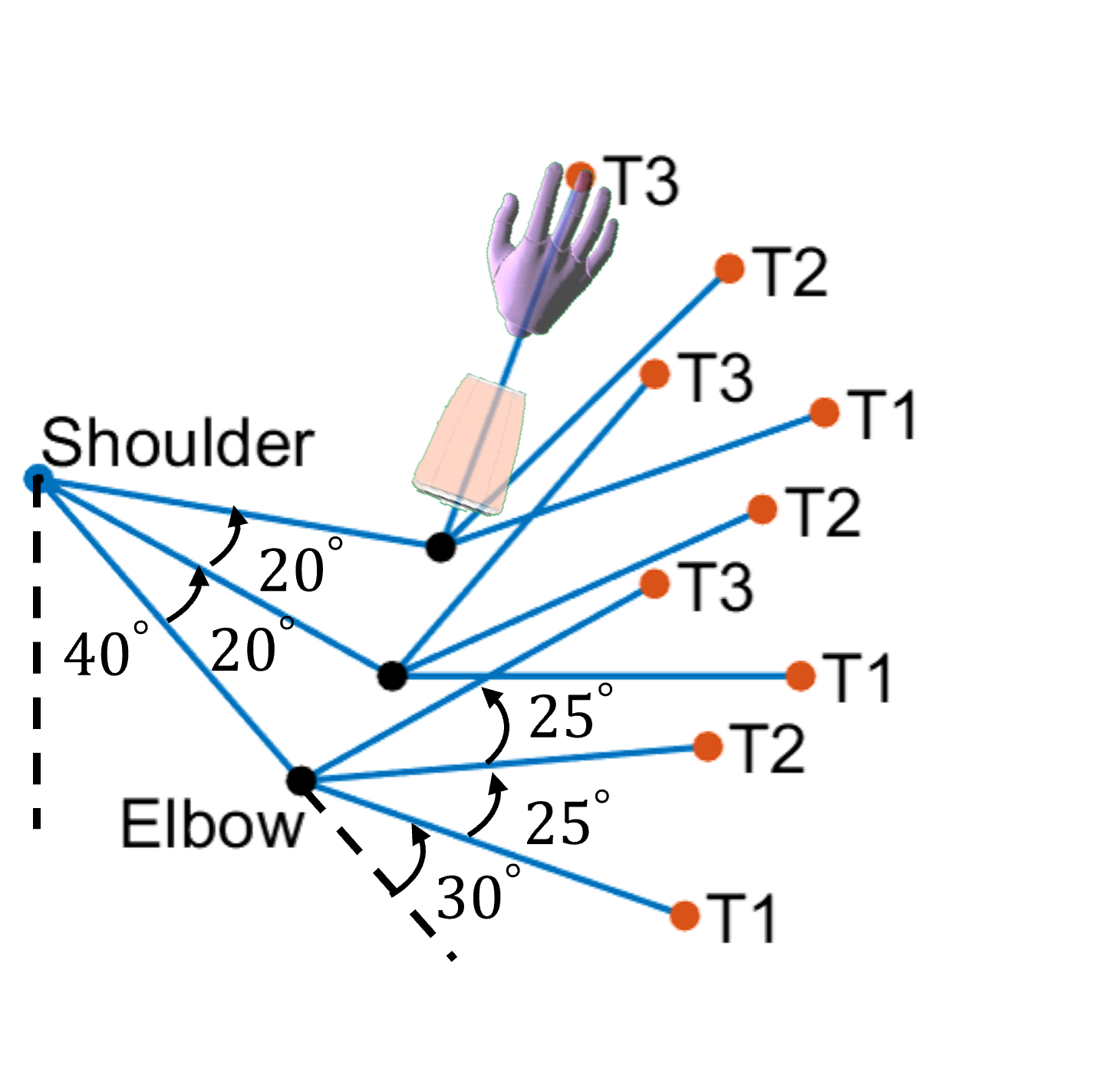}
\label{fig:prost_setup:a}}%
\subfloat[]{%
\includegraphics[width=0.27\textwidth]{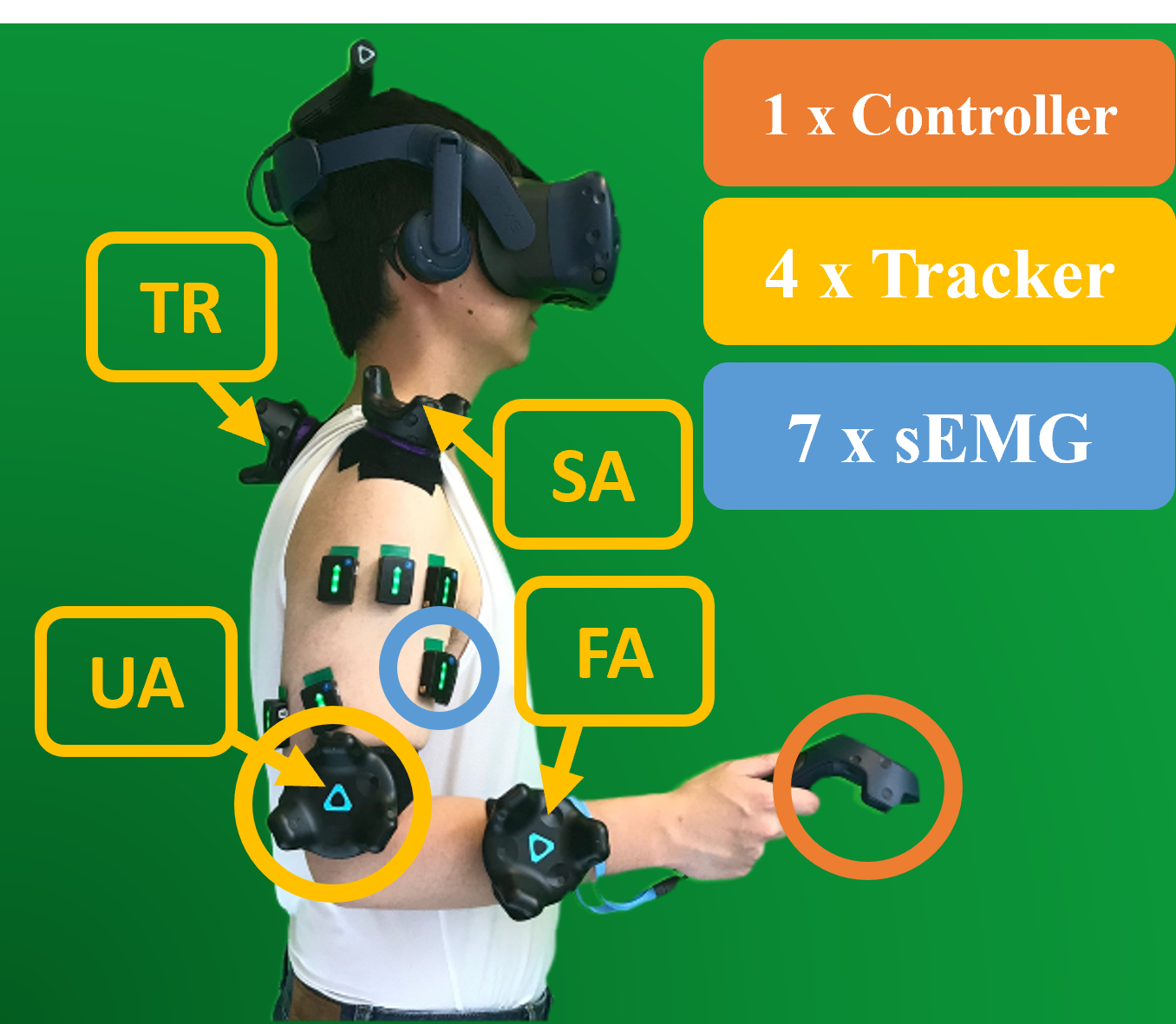}
\label{fig:prost_setup:b}}%
\hspace{12mm}
\subfloat[]{%
\includegraphics[width=0.27\textwidth]{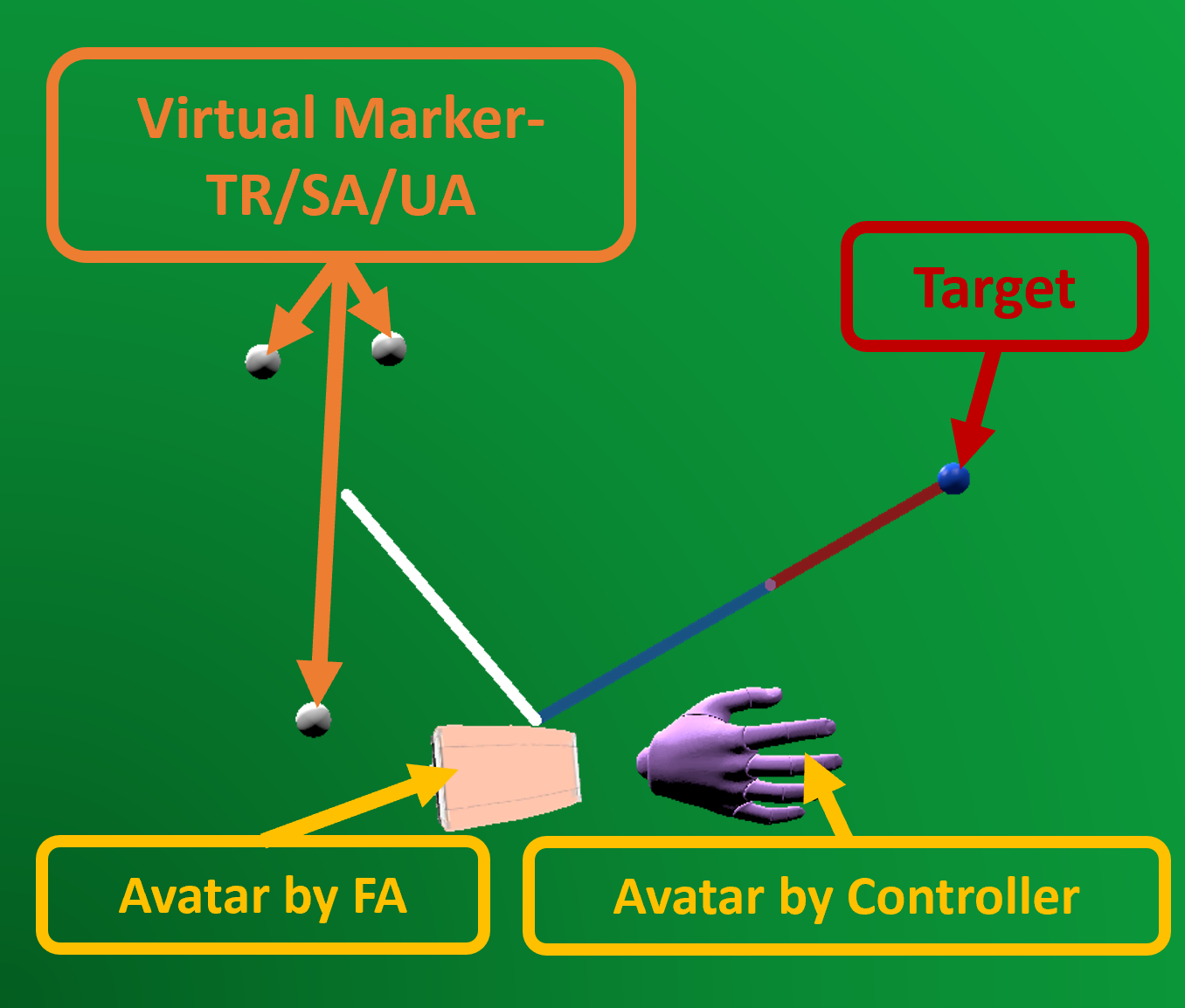}
\label{fig:prost_setup:c}}%
\hspace{12mm}

\caption{(a) Target shoulder and elbow poses, T1--T3 denote the three goals, (b) Experimental setup and the placement of VIVE trackers and sEMG electrodes, (c) VR avatar and reaching target example (side view). }
\label{fig:prost_setup}
\vspace{-4mm}
\end{figure*}
     
\subsection{Methods of Data Collection}
    This subsection elaborates on the dataset and data collection procedure. Ten healthy subjects (seven males and three females), all right-handed, were recruited. The experimental protocol was approved by the University of Melbourne Human Research Ethics Committee under project ID 11878. Informed consent was obtained from all the subjects.

    The dataset contains 35 sEMG and 12 joint kinematic movement features of the subjects intending for three goals at three different shoulder poses. The features were sampled at a rate of 10 Hz (the measurements were taken every 0.1 seconds). The goals (target elbow poses) are denoted as $\text{T1}$, $\text{T2}$, $\text{T3}$ in \cref{fig:prost_setup:a} which shows the side-view schematic of the upper limb. The process of extracting the sEMG and joint pose features has been described in detail in our previous work~\cite{Prost:Yu2023}.

    The sensor setup and virtual avatar in VR are shown in \cref{fig:prost_setup:b} and \cref{fig:prost_setup:c}, respectively. In the experiment, the subjects stood naturally and unconstrained in an upright position to perform forward-reaching tasks by extending their upper limb towards the three goals at three shoulder poses. The goal is considered reached when the middle finger of the virtual avatar hits the target sphere, generated to elicit the target shoulder and elbow poses as shown in \cref{fig:prost_setup:c}. For each goal, the subjects completed 30 iterations of reaching. During each iteration, the subjects were instructed to maintain their final upper limb pose for one second upon reaching the goal. The recorded data spanned from the start of the movement to the end of the holding period. 
    
    Given the focus on transhumeral prostheses, the kinematic movements of the above-elbow joints were recorded along with the muscle activity of the upper arm. To capture the kinematics, three HTC VIVE trackers were strategically positioned at the upper arm (UA), shoulder acromion (SA), and trunk (TR). An additional tracker was placed on the forearm (FA), and a controller was held in the hand solely for controlling the forearm and hand avatar within the VR environment, as shown in \cref{fig:prost_setup:c}. For monitoring the muscle activity, seven Delsys\textsuperscript{\textregistered} Trigno\textsuperscript{\texttrademark} wireless sEMG electrodes were attached to the muscles of the dominant upper arm of each subject.

%% file: tex/related_work.tex
\section{Related Work}
\label{sec:related:work}

Existing GR methods can be broadly classified into three categories: plan library~\cite{DBLP:conf/aaai/KautzA86}, planning-based~\cite{DBLP:conf/aaai/RamirezG10}, and data-driven~\cite{DBLP:conf/ijcai/MinMRLL16,DBLP:conf/atal/PolyvyanyySLS20,AIJ_Su} methods.
A plan library GR method relies on exemplary plans for accomplishing candidate goals, usually hand-crafted by domain experts.
The method proceeds by comparing the observed agent actions with the available example plans to infer the true goal of the agent.
The planning-based GR techniques use pre-defined domain models that describe the environment and possible actions agents can take in the environment.
Once a new observed sequence of actions in the environment is available, these techniques rely on the domain models to construct and compare plans toward the candidate goals to infer the likelihoods of the goals.
Finally, the data-driven techniques use historical data collected from agents and the environment to perform goal inference.

In the field of transhumeral prostheses, previous studies have demonstrated that employing diverse features tailored to individual patients results in improved performance when classifying target poses~\cite{DBLP:conf/smc/YuG00CO21}.
This personalized approach to handling patients complicates crafting exemplary plans and domain models, as each patient requires dedicated plans and models.
In this light, data-driven GR techniques appear appropriate for the individual user, as they can learn personalized artifacts from the historical behavior of a patient for subsequent goal inferences tuned for the patient.
In this work, we apply the data-driven GR at the level of individual subjects.

The data-driven GR approach based on LSTM neural networks~\cite{DBLP:conf/ijcai/MinMRLL16} learns a model from a set of action sequences.
It then identifies the most likely goal of the agent based on a newly observed sequence of actions.
Based on a collection of historical sequences of actions towards a candidate goal, the PM-based GR system~\cite{DBLP:conf/atal/PolyvyanyySLS20,AIJ_Su} uses process discovery techniques to learn a process model that describes the skill for accomplishing the goal. 
Subsequently, it relies on conformance checking techniques~\cite{DBLP:journals/widm/AalstAD12} to study commonalities and discrepancies between a newly observed sequence of actions and all the learned models, one model for each candidate goal. 
These commonalities and discrepancies are then translated into a probability distribution over the candidate goals, capturing the likelihoods that the actions aim to reach these goals.

Machine learning techniques can be used to implement accurate prosthesis control~\cite{Prost:Shehata2021}.
Due to their robustness, machine learning classifiers are commonly implemented in upper-limb prostheses~\cite{Geethanjali2016}. 
Given an input signal, for example, from sensors, a classifier predicts the output signal and, the intended movement of the patient.
Linear discriminant analysis (LDA) is the most commonly used classifier algorithm in prosthesis control.
It can provide high control accuracy using small training and processing times~\cite{Parajuli2019}.
Our recent work~\cite{DBLP:conf/smc/YuG00CO21} confirmed that LDA can discriminate poses reliably.
Recent works explore the use of artificial neural networks in prosthesis control~\cite{Prost:Shehata2021}.
Neural networks can achieve high goal recognition accuracy by learning non-linear dependencies between signal input and control outputs but require extensive training over large datasets~\cite{Prost:Shehata2021}.
For instance, Huang et al.~\cite{huang2021development} successfully used LSTM neural networks to predict target poses based on time series of electromyography signals.

This paper aims to understand how the PM-based GR system can contribute to developing active transhumeral prostheses.
To this end, we extend the system to work with repeated, real-valued, continuous data captured by typical sensors and propose dedicated training techniques.
We use LDA~\cite{DBLP:conf/smc/YuG00CO21} and LSTM~\cite{DBLP:conf/ijcai/MinMRLL16,huang2021development} models as performance baselines.

%% file: tex/approach.tex
\section{Approach}
\label{sec:approach}
This section presents the PM-based GR system~\cite{DBLP:conf/atal/PolyvyanyySLS20,AIJ_Su} and its new training approach based on real-valued, continuous data.
The approach starts by reducing the data dimensionality by selecting relevant features (cf. \cref{subsec:feature:selection}).
Next, the data points of reduced dimensionality are clustered to define discrete events (cf. \cref{subsec:event:discretization}).
\Cref{subsec:discovery} explains our approach to discover process models from discrete events.
Finally, \cref{subsec:recognition} summarizes our technique for goal inference based on conformance diagnostics between the discovered models and new observations.

To illustrate the approach, we use six traces in which events are characterized by 30 real-valued, continuous features $f_1$ to $f_{30}$; the dataset and tool for replicating the running example are publicly available at \href{https://doi.org/10.26188/24131493}{https://doi.org/10.26188/24131493}.
The first three traces (1--3) represent the signal sequences collected when reaching the target pose T1, while the last three traces (4--6) were recorded when reaching the target pose T2.
\Cref{tab:src_example} shows an extract of the dataset. 
In the table, each row holds feature values that were collected simultaneously and the rows are ordered by their time stamp of data collection. 

\input{tables/src_example.tex}

\subsection{Feature Selection}
\label{subsec:feature:selection}

\begin{figure}[t]
\vspace{1mm}
\centering
\includegraphics[width=0.45\textwidth]{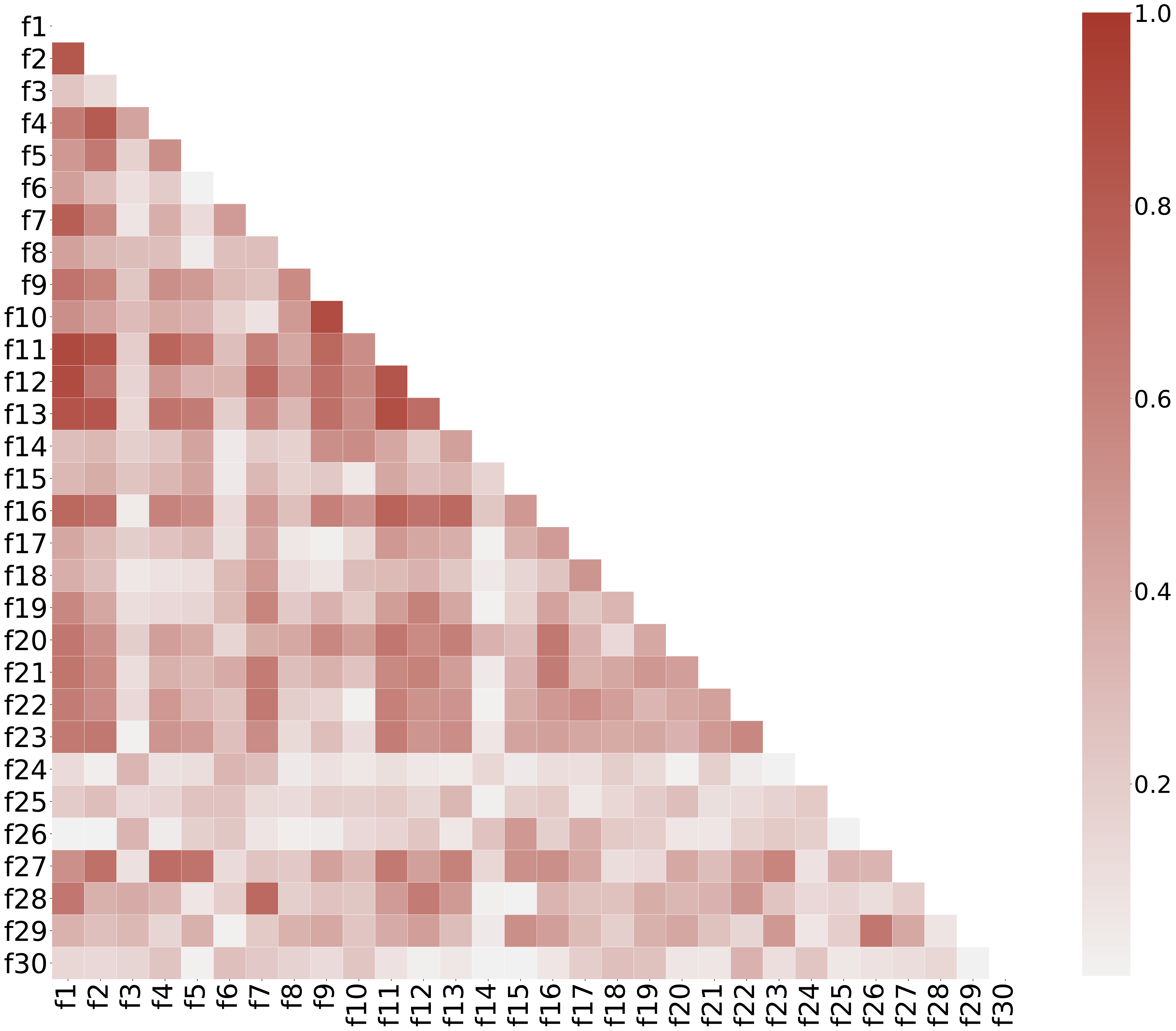}
\caption{Correlation matrix.}
\label{fig:corr}
\vspace{-1mm}
\end{figure}

\begin{figure}[ht!]
\vspace{-2mm}
\centering
\includegraphics[width=0.45\textwidth]{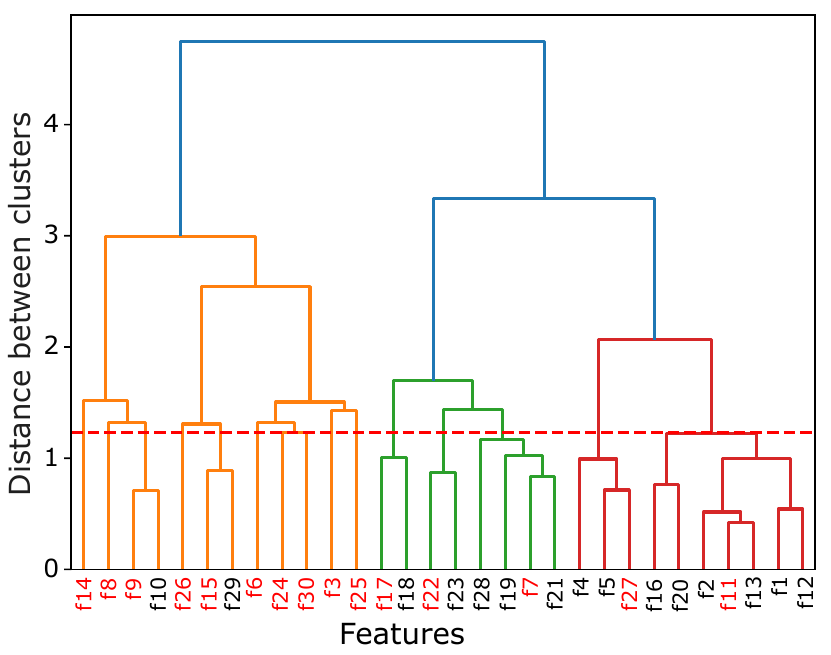}
\caption{Dendrogram and selection of clusters. The red dotted line represents the threshold used to cut the dendrogram to form 15 clusters.}
\label{fig:dend}
\vspace{-2mm}
\end{figure}

To reduce the dimensionality of the data, we eliminate highly correlated features from the analysis.
We use the agglomerative hierarchical clustering method~\cite{day1984efficient} to cluster the features into $N_f$ groups based on the distances defined by the correlations between the features.
\Cref{fig:corr} shows the matrix of absolute values of the Pearson correlation between the features, while~\cref{fig:dend} depicts the dendrogram of the clusters.
The dendrogram represents the hierarchical structure of clusters, which supports the flexible selection of the desired features using a similarity threshold.
A similarity threshold value defines the desired distance between the constructed clusters.
For instance, \cref{fig:dend} shows the selection of 15 clusters using the similarity threshold value of 1.23.
A cluster is defined by all the leaf features in a dendrogram branch cut horizontally by the threshold line.
The systematic approach for selecting the features is discussed in \cref{sec:evaluation}. 
Here, we use $N_f$ of 15 for demonstration.

As the features within a cluster are highly correlated, we select the medoid of the cluster, the feature with the smallest average dissimilarity to all other features in the cluster, as the representative of the cluster.
This way, we obtain $N_f$ features that characterize the source data.
In the running example, we reduce the original 30 features to ${N_f=15}$ representative features highlighted in \cref{fig:dend} in red and listed in \cref{tab:reduced_example}.

\input{tables/reduced_example.tex}

\subsection{Event Discretization}
\label{subsec:event:discretization}

We transform the $N_f$-dimensional time series of real-valued, continuous measurements obtained in the previous step into traces of events amenable to process mining.
To this end, we cluster the $N_f$-dimensional data points from the series into $N_c$ clusters.
The data points in a cluster are accepted as the same event.
We use $k$-means algorithm~\cite{hartigan1979algorithm} due to the performance considerations for clustering high-dimensional data.

In the example, we clustered the 15-dimensional data points defined by the 15 selected in the previous step features into 10 clusters, represented by events $e_0$ to $e_9$, see the ``Event'' column in \cref{tab:reduced_example}. 
Again, the systematic approach for selecting the number of clusters is discussed in \cref{sec:evaluation}. 
Here, we use $N_c = 10$ for demonstration only.
The traces of obtained events can be split into two event logs, containing the traces towards goals T1 and T2, denoted by $L_1$ and $L_2$, respectively.

\subsection{Process Discovery}
\label{subsec:discovery}

\begin{figure}[h!]
\vspace{-8pt}
\centering
\scalebox{0.4}{\input{fig/tikz/pn2_clean.tex}}
\vspace{-2pt}
\caption{Process model $M_2$ discovered from even log $L_2$.}
\label{fig:pn2}
\vspace{-2pt}
\end{figure}
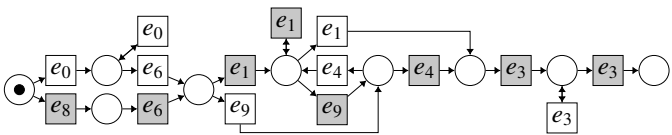

Given an event log of discrete events towards a target pose, we use process discovery to construct a process model that describes and generalizes the historical traces that were followed in the past to achieve the corresponding goal.
For example, we use the Directly Follows Miner algorithm~\cite{DBLP:conf/icpm/LeemansPW19} to construct Petri nets $M_1$ and $M_2$ shown in \cref{fig:pn1,fig:pn2} from event logs $L_1$ and $L_2$ obtained in the previous step.

Since event logs $L_1$ and $L_2$ solely consist of traces related to reaching target poses T1 and T2, respectively, process models $M_1$ and $M_2$ depict the potential processes involved in reaching corresponding final target poses T1 and T2.
Models $M_1$ and $M_2$ are stored in our GR system as ``knowledge,'' which assists in predicting the target pose of the prosthesis when the GR system receives new sequences of signals.

\begin{figure}[h!]
\centering
\scalebox{0.4}{\input{fig/tikz/pn1_clean.tex}}
\vspace{-2pt}
\caption{Process model $M_1$ discovered from even log $L_1$.}
\label{fig:pn1}
\end{figure}
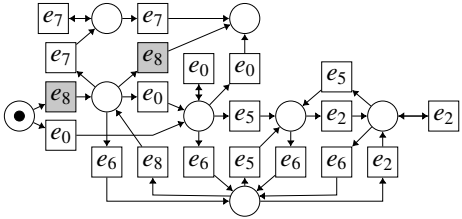

\subsection{Goal Recognition}
\label{subsec:recognition}

Once a new (prefix of a) trace of multi-dimensional, real-valued, continuous measurements is observed, it is used to infer the goal of the patient.
The trace is reduced to the representative features and transformed into an event trace.
The selection of representative features is described in \cref{subsec:feature:selection}.
A data point is attributed to the closest event cluster out of the clusters obtained as described in \cref{subsec:event:discretization}.
For example, suppose we observe this new trace of events:
$$\tau = \left\langle e_8, e_6, e_2, e_1, e_1, e_9 \right\rangle.$$

\noindent
The PM-based GR system uses conformance checking to diagnose discrepancies and commonalities between the event trace and all the discovered process models.
Specifically, it constructs optimal alignments~\cite{DBLP:journals/widm/AalstAD12} between the trace and the models.
Below, we show optimal alignments $\sigma_1$ and $\sigma_2$ between example trace $\tau$ and process models $M_1$ and $M_2$.
The transitions of Petri nets $M_1$ and $M_2$ that participate in the alignments with $\tau$ are highlighted in gray in the figures.
\input{tables/alignments.tex}

\noindent
An alignment is a sequence of moves.
In a table that encodes an alignment, columns represent moves.
A move with the special ``$\gg$'' skip symbol is an asynchronous move.
Otherwise, a move is a synchronous move.
A synchronous move represents that both the trace and model can proceed synchronously by executing the same action.
A move with the skip in the bottom (top) row is a move on trace (on model), denoting that the model (trace) cannot mimic the action in the trace (model).
An optimal alignment is an alignment with the ``cheapest'' asynchronous moves according to some predefined positive costs of asynchronous moves that capture the minimal (as per the costs) deviations between the trace and model.
For example, optimal alignment $\sigma_1$ starts with a synchronous move, denoting that both the trace and the model can start with event $e_8$, followed by five asynchronous moves on trace, capturing that it is cheaper for the model not to match these occurrences of events in the trace.

The more synchronous moves between trace $\tau$ and model $M_G$ in the alignment, the higher the likelihood that the trace is following $M_G$, indicating an intention to reach goal $G$. 
In the running example, observed trace $\tau$ is more likely to reach target pose T2.
The detailed probability calculation uses \emph{alignment weight} between $\tau$ and $M_G$, as defined below~\cite{DBLP:conf/atal/PolyvyanyySLS20}.

\begin{equation}
	\omega(\tau,M_G) = \phi + \lambda^m  \times \sum\limits_{i=1}^{n} \left( i^{\,\delta} \times c(\tau,M_G,i) \right)
	\label{eq:omega}
\end{equation}

\noindent
In~\cref{eq:omega},
$c(\tau,M_G,i)$ is the cost of the move in the alignment between $\tau$ and $M_G$ at position $i$ (we use a cost of one for all asynchronous moves on trace and a cost of zero for all other moves),
$n$ is the length of the alignment,
$\phi$ is a constant,
$\delta$ is a discount factor that emphasizes later asynchronous moves in the alignment,
$\lambda \geq 1$ is a penalty for asynchronous moves on trace, and
$m$ is the number of consecutive asynchronous moves on trace occurring at the end of the alignment.
Intuitively, the more asynchronous moves at the end of the alignment, the greater the alignment weight.

The alignment weight is used to compute the probability of reaching the corresponding goal, as captured below~\cite{DBLP:conf/atal/PolyvyanyySLS20}.

\begin{equation}\label{eq:probability}
	\Pr(G \mid \tau)  =  
		\frac{e^{-\beta \,\times\, \omega(\tau,M_G)}}
			{\sum\limits_{G' \in \mathcal{G}} e^{-\beta \,\times\, \omega(\tau,M_{G'})}}
\end{equation} 

\noindent
In~\cref{eq:probability}, $\mathcal{G}$ is the set of candidate goals, $G'$ denotes each goal candidate in the set $\mathcal{G}$, and $\beta \in (0,1]$ is the level of trust over the discovered process models.
Based on the two example alignments, the probabilities of trace $\tau$ reaching T1 and T2 are 0.06 and 0.94, respectively, using the default parameter setting~\cite{DBLP:conf/atal/PolyvyanyySLS20}.
Consequently, the PM-based GR system infers that trace $\tau$ intends to reach T2.


%% file: tables/src_example.tex
\begin{table}[h]
\vspace{-1mm}
\centering
\resizebox{0.48\textwidth}{!}{%
\begin{tabular}{|c|c|c|c|c|c|c|c|}
\hline
Trace & Goal & $f_1$         & $f_2$         & $f_3$         & \ldots & $f_{29}$        & $f_{30}$ \\ 
\hline
\hline
1        & T1    & 5.19727337 & 7.02395793 & 0.00254431 & \ldots & 5.39759498 & -0.3722619 \\ \hline
1        & T1    & 7.76278776 & 8.08816201 & 0.00472689 & \ldots & 1.01557531 & 1.37592798 \\ \hline
1        & T1    & 13.4185557 & 8.87159453 & 0.00821896 & \ldots & -4.0004147 & 1.65328609 \\ \hline
1        & T1    & 22.0916619 & 9.04377674 & 0.01015369 & \ldots & -5.5399488 & -1.7805512 \\ \hline
1        & T1    & 31.3641039 & 9.3586209  & 0.009165   & \ldots & -3.5156837 & 1.36367015 \\ \hline
1        & T1    & 38.2312577 & 10.139119  & 0.00616715 & \ldots & -1.4720033 & 5.87820456 \\ \hline
1        & T1    & 42.0592085 & 10.8827908 & 0.00315491 & \ldots & -0.3338844 & 4.29640897 \\ \hline
2        & T1    & 7.39110795 & 6.07336937 & 0.00064332 & \ldots & 2.92403705 & 1.46698529 \\ \hline
2        & T1    & 10.5229866 & 7.44734189 & 0.00194998 & \ldots & 1.60034347 & 2.94734496 \\ \hline
\ldots      & \ldots  & \ldots        & \ldots        & \ldots        & \ldots & \ldots        & \ldots        \\ \hline
6        & T2    & 64.1830578 & 25.2975943 & -0.0003433 & \ldots & -1.1970367 & 0.92412363 \\ \hline
6        & T2    & 66.8916142 & 27.5304609 & -0.0017204 & \ldots & 0.31022101 & 0.95595258 \\ \hline
\end{tabular}%
}
\caption{Extract of the running example dataset.}
\label{tab:src_example}
\vspace{-2mm}
\end{table}

%% file: tables/reduced_example.tex
\begin{table}[h]
\vspace{3mm}
\centering
\resizebox{0.48\textwidth}{!}{%
\begin{tabular}{|c|c|l|l|c|l|l||c|}
\hline
Trace & Goal & \multicolumn{1}{c|}{$f_{3}$}  & \multicolumn{1}{c|}{$f_{6}$}  & \ldots & \multicolumn{1}{c|}{$f_{27}$} & \multicolumn{1}{c||}{$f_{30}$} & Event \\ 
\hline
\hline
1 & T1 & 0.002544311 & 0.121301538 & \ldots & 9.057075924 & -0.372261852 & $e_0$ \\ \hline
1 & T1 & 0.004726894 & 0.210557727 & \ldots & 10.55266625 & 1.375927982 & $e_6$ \\ \hline
1 & T1 & 0.008218956 & 0.391712037 & \ldots & 4.416704571 & 1.653286092 & $e_5$ \\ \hline
1 & T1 & 0.010153689 & 0.327711654 & \ldots & 1.372325318 & -1.780551234 & $e_2$ \\ \hline
1 & T1 & 0.009165 & 0.311548058 & \ldots & 5.526959903 & 1.363670147 & $e_5$ \\ \hline
1 & T1 & 0.00616715 & 0.734098175 & \ldots & 8.869853729 & 5.87820456 & $e_6$ \\ \hline
1 & T1 & 0.003154906 & 1.227944362 & \ldots & 4.916643199 & 4.296408971 & $e_8$ \\ \hline
2 & T1 & 0.000643317 & 0.103164637 & \ldots & 12.86040763 & 1.466985286 & $e_8$ \\ \hline
2 & T1 & 0.001949978 & 0.336630569 & \ldots & 13.45204634 & 2.94734496 & $e_6$ \\ \hline
\ldots      & \ldots  & \multicolumn{1}{c|}{\ldots} & \multicolumn{1}{c|}{\ldots} & \ldots & \multicolumn{1}{c|}{\ldots} & \multicolumn{1}{c||}{\ldots} & {\ldots}            \\ \hline
6 & T2 & -0.000343322 & 0.410761081 & \ldots & 23.78469914 & 0.924123631 & $e_4$ \\ \hline
6 & T2 & -0.001720367 & 0.504684583 & \ldots & 21.08908217 & 0.955952575 & $e_3$ \\ \hline
\end{tabular}%
}
\caption{Extract of the reduced running example dataset.}
\label{tab:reduced_example}
\vspace{-3mm}
\end{table}

%% file: fig/tikz/pn1_clean.tex
\begin{tikzpicture}[thick,font=\Huge]

\newcommand{\hshift}{0.5cm}
\newcommand{\vshift}{0.275cm}
\definecolor{lightgray}{RGB}{200,200,200} 

\node[draw,align=center,circle,minimum size=1cm] (p1) at (0,0) {};
\node[draw,fill,circle,minimum size=3mm] (token) at (p1.center) {};

\node[xshift=\hshift,yshift=\vshift,draw,fill=lightgray,align=center,minimum size=1cm,anchor=west] (t1) at (p1.north east) {$e_8$};
\node[xshift=\hshift,yshift=-\vshift,draw,align=center,minimum size=1cm,anchor=west] (t2) at (p1.south east) {$e_0$};

\node[xshift=\hshift,draw,align=center,circle,minimum size=1cm,anchor=west] (p2) at (t1.east) {};

\node[xshift=\hshift,draw,align=center,minimum size=1cm,anchor=west] (t3) at (p2.east) {$e_0$};

\node[xshift=\hshift,draw,align=center,circle,minimum size=1cm,anchor=west] (p3) at (t3.east |- p1) {};

\node[yshift=0.5*\vshift,draw,align=center,minimum size=1cm,anchor=north] (t4) at (p2.south |- t2.south) {$e_6$};
\node[xshift=\hshift,draw,align=center,minimum size=1cm,anchor=west] (t5) at (t4.east) {$e_8$};
\node[xshift=\hshift,draw,align=center,minimum size=1cm,anchor=west] (t6) at (t5.east) {$e_6$};
\node[xshift=\hshift,draw,align=center,minimum size=1cm,anchor=west] (t7) at (t6.east) {$e_5$};
\node[xshift=\hshift,draw,align=center,minimum size=1cm,anchor=west] (t8) at (t7.east) {$e_6$};
\node[xshift=\hshift,draw,align=center,minimum size=1cm,anchor=west] (t9) at (t8.east) {$e_6$};
\node[xshift=\hshift,draw,align=center,minimum size=1cm,anchor=west] (t10) at (t9.east) {$e_2$};

\node[yshift=-\vshift,draw,align=center,circle,minimum size=1cm,anchor=north] (p4) at (t7.south) {};

\node[yshift=2*\vshift,draw,align=center,minimum size=1cm,anchor=south] (t11) at (p3.north) {$e_0$};

\node[xshift=\hshift,draw,align=center,minimum size=1cm,anchor=west] (t12) at (p3.east) {$e_5$};
\node[xshift=\hshift,draw,align=center,circle,minimum size=1cm,anchor=west] (p5) at (t12.east) {};
\node[xshift=\hshift,draw,align=center,minimum size=1cm,anchor=west] (t13) at (p5.east) {$e_2$};
\node[xshift=\hshift,draw,align=center,circle,minimum size=1cm,anchor=west] (p6) at (t13.east) {};
\node[xshift=2*\hshift,draw,align=center,minimum size=1cm,anchor=west] (t14) at (p6.east) {$e_2$};

\node[yshift=\vshift,,draw,align=center,minimum size=1cm,anchor=south] (t15) at (t13.north) {$e_5$};

\node[yshift=\vshift,draw,fill=lightgray,align=center,minimum size=1cm,anchor=south] (t16) at (t3.north) {$e_8$};

\node[draw,align=center,minimum size=1cm] (t17) at (t12 |- t11) {$e_0$};

\node[draw,align=center,minimum size=1cm] (t18) at (t2 |- t16) {$e_7$};
\node[yshift=\vshift,draw,align=center,circle,minimum size=1cm,anchor=south] (p7) at (p2 |- t18.north) {};
\node[yshift=\vshift,draw,align=center,minimum size=1cm,anchor=south] (t19) at (t16.north) {$e_7$};

\node[xshift=-1.5*\hshift,draw,align=center,minimum size=1cm,anchor=east] (t20) at (p7.west) {$e_7$};

\node[draw,align=center,circle,minimum size=1cm] (final) at (t12 |- t19) {};

\draw[-{Latex[length=2mm,width=2.5mm]}] (p1) -- (t1);
\draw[-{Latex[length=2mm,width=2.5mm]}] (p1) -- (t2);
\draw[-{Latex[length=2mm,width=2.5mm]}] (t1) -- (p2);
\draw[-{Latex[length=2mm,width=2.5mm]}] (t2) -- (t3 |- t2) -- (p3);

\draw[-{Latex[length=2mm,width=2.5mm]}] (p2) -- (t3);
\draw[-{Latex[length=2mm,width=2.5mm]}] (t3) -- (p3);

\draw[-{Latex[length=2mm,width=2.5mm]}] (p2) -- (t4);
\draw[-{Latex[length=2mm,width=2.5mm]}] (t5) -- (p2);

\draw[-{Latex[length=2mm,width=2.5mm]}] (t4) -- (t4 |- p4) -- (p4);
\draw[-{Latex[length=2mm,width=2.5mm]}] ($(p4.center) + (160:0.5cm)$) -- ([yshift=0.75*\vshift]t5 |- p4) -- (t5);
\draw[-{Latex[length=2mm,width=2.5mm]}] (t6) -- (p4);
\draw[-{Latex[length=2mm,width=2.5mm]}] (p4) -- (t7);
\draw[-{Latex[length=2mm,width=2.5mm]}] (t8) -- (p4);
\draw[-{Latex[length=2mm,width=2.5mm]}] (t9) -- ([yshift=0.75*\vshift]t9 |- p4) -- ($(p4.center) + (20:0.5cm)$);
\draw[-{Latex[length=2mm,width=2.5mm]}] (p4) -- (t10 |- p4) -- (t10);
\draw[-{Latex[length=2mm,width=2.5mm]}] (p3) -- (t6);

\draw[-{Latex[length=2mm,width=2.5mm]}] (p3) -- (t11);
\draw[-{Latex[length=2mm,width=2.5mm]}] (t11) -- (p3);

\draw[-{Latex[length=2mm,width=2.5mm]}] (p3) -- (t12);
\draw[-{Latex[length=2mm,width=2.5mm]}] (t12) -- (p5);
\draw[-{Latex[length=2mm,width=2.5mm]}] (p5) -- (t13);
\draw[-{Latex[length=2mm,width=2.5mm]}] (t13) -- (p6);
\draw[-{Latex[length=2mm,width=2.5mm]}] (p6) -- (t14);
\draw[-{Latex[length=2mm,width=2.5mm]}] (t14) -- (p6);

\draw[-{Latex[length=2mm,width=2.5mm]}] (t7) -- (p5);
\draw[-{Latex[length=2mm,width=2.5mm]}] (p5) -- (t8);

\draw[-{Latex[length=2mm,width=2.5mm]}] (p6) -- (t9);
\draw[-{Latex[length=2mm,width=2.5mm]}] (t10) -- (p6);

\draw[-{Latex[length=2mm,width=2.5mm]}] (p6) -- (t15);
\draw[-{Latex[length=2mm,width=2.5mm]}] (t15) -- (p5);

\draw[-{Latex[length=2mm,width=2.5mm]}] (p2) -- (t16);
\draw[-{Latex[length=2mm,width=2.5mm]}] (t16) -- (final);

\draw[-{Latex[length=2mm,width=2.5mm]}] (p3) -- (t17);
\draw[-{Latex[length=2mm,width=2.5mm]}] (t17) -- (final);

\draw[-{Latex[length=2mm,width=2.5mm]}] (p2) -- (t18);
\draw[-{Latex[length=2mm,width=2.5mm]}] (t18) -- (p7);
\draw[-{Latex[length=2mm,width=2.5mm]}] (p7) -- (t19);
\draw[-{Latex[length=2mm,width=2.5mm]}] (t19) -- (final);

\draw[-{Latex[length=2mm,width=2.5mm]}] (p7) -- (t20);
\draw[-{Latex[length=2mm,width=2.5mm]}] (t20) -- (p7);

\end{tikzpicture}

%% file: fig/tikz/pn2_clean.tex
\begin{tikzpicture}[thick,font=\Huge]

\newcommand{\hshift}{0.5cm}
\newcommand{\vshift}{0.275cm}
\definecolor{lightgray}{RGB}{200,200,200} 

\node[draw,align=center,circle,minimum size=1cm] (p1) at (0,0) {};
\node[draw,fill,circle,minimum size=3mm] (token) at (p1.center) {};

\node[xshift=\hshift,yshift=\vshift,draw,align=center,minimum size=1cm,anchor=west] (t1) at (p1.north east) {$e_0$};
\node[xshift=\hshift,yshift=-\vshift,draw,fill=lightgray,align=center,minimum size=1cm,anchor=west] (t2) at (p1.south east) {$e_8$};

\node[xshift=\hshift,draw,align=center,circle,minimum size=1cm,anchor=west] (p2) at (t1.east) {};
\node[xshift=\hshift,draw,align=center,circle,minimum size=1cm,anchor=west] (p3) at (t2.east) {};

\node[xshift=\hshift,draw,align=center,minimum size=1cm,anchor=west] (t3) at (p2.east) {$e_6$};
\node[xshift=\hshift,draw,fill=lightgray,align=center,minimum size=1cm,anchor=west] (t4) at (p3.east) {$e_6$};

\node[xshift=\hshift,draw,align=center,circle,minimum size=1cm,anchor=west] (p4) at (t3.east |- p1) {};

\node[yshift=\vshift,draw,align=center,minimum size=1cm,anchor=south] (t5) at (t3.north) {$e_0$};

\node[xshift=\hshift,yshift=\vshift,draw,fill=lightgray,align=center,minimum size=1cm,anchor=west] (t6) at (p4.north east) {$e_1$};
\node[xshift=\hshift,yshift=-\vshift,draw,align=center,minimum size=1cm,anchor=west] (t7) at (p4.south east) {$e_9$};

\node[xshift=\hshift,draw,align=center,circle,minimum size=1cm,anchor=west] (p5) at (t6.east) {};

\node[xshift=\hshift,draw,align=center,minimum size=1cm,anchor=west] (t9) at (p5.east) {$e_4$};
\node[yshift=\vshift,draw,align=center,minimum size=1cm,anchor=south] (t8) at (t9.north) {$e_1$};
\node[yshift=-\vshift,draw,fill=lightgray,align=center,minimum size=1cm,anchor=north] (t10) at (t9.south) {$e_9$};

\node[yshift=2*\vshift,draw,fill=lightgray,align=center,minimum size=1cm,anchor=south] (t11) at (p5.north) {$e_1$};

\node[xshift=\hshift,draw,align=center,circle,minimum size=1cm,anchor=west] (p6) at (t9.east) {};

\node[xshift=\hshift,draw,fill=lightgray,align=center,minimum size=1cm,anchor=west] (t12) at (p6.east) {$e_4$};
\node[xshift=\hshift,draw,align=center,circle,minimum size=1cm,anchor=west] (p7) at (t12.east) {};
\node[xshift=\hshift,draw,fill=lightgray,align=center,minimum size=1cm,anchor=west] (t13) at (p7.east) {$e_3$};
\node[xshift=\hshift,draw,align=center,circle,minimum size=1cm,anchor=west] (p8) at (t13.east) {};

\node[xshift=\hshift,draw,fill=lightgray,align=center,minimum size=1cm,anchor=west] (t14) at (p8.east) {$e_3$};
\node[xshift=\hshift,draw,align=center,circle,minimum size=1cm,anchor=west] (p9) at (t14.east) {};

\node[yshift=-2*\vshift,draw,align=center,minimum size=1cm,anchor=north] (t15) at (p8.south) {$e_3$};

\draw[-{Latex[length=2mm,width=2.5mm]}] (p1) -- (t1);
\draw[-{Latex[length=2mm,width=2.5mm]}] (p1) -- (t2);
\draw[-{Latex[length=2mm,width=2.5mm]}] (t1) -- (p2);
\draw[-{Latex[length=2mm,width=2.5mm]}] (t2) -- (p3);

\draw[-{Latex[length=2mm,width=2.5mm]}] (p2) -- (t5);
\draw[-{Latex[length=2mm,width=2.5mm]}] (t5) -- (p2);

\draw[-{Latex[length=2mm,width=2.5mm]}] (p2) -- (t3);
\draw[-{Latex[length=2mm,width=2.5mm]}] (p3) -- (t4);
\draw[-{Latex[length=2mm,width=2.5mm]}] (t3) -- (p4);
\draw[-{Latex[length=2mm,width=2.5mm]}] (t4) -- (p4);

\draw[-{Latex[length=2mm,width=2.5mm]}] (p4) -- (t6);
\draw[-{Latex[length=2mm,width=2.5mm]}] (p4) -- (t7);

\draw[-{Latex[length=2mm,width=2.5mm]}] (t6) -- (p5);
\draw[-{Latex[length=2mm,width=2.5mm]}] (p5) -- (t8);
\draw[-{Latex[length=2mm,width=2.5mm]}] (t9) -- (p5);
\draw[-{Latex[length=2mm,width=2.5mm]}] (p5) -- (t10);

\draw[-{Latex[length=2mm,width=2.5mm]}] (p5) -- (t11);
\draw[-{Latex[length=2mm,width=2.5mm]}] (t11) -- (p5);

\draw[-{Latex[length=2mm,width=2.5mm]}] (p6) -- (t9);
\draw[-{Latex[length=2mm,width=2.5mm]}] (t10) -- (p6);

\draw[-{Latex[length=2mm,width=2.5mm]}] (t7) -- ([yshift=-\vshift]t7.south) -- ([yshift=-\vshift]p6 |- t7.south) -- (p6);

\draw[-{Latex[length=2mm,width=2.5mm]}] (p6) -- (t12);
\draw[-{Latex[length=2mm,width=2.5mm]}] (t12) -- (p7);
\draw[-{Latex[length=2mm,width=2.5mm]}] (p7) -- (t13);
\draw[-{Latex[length=2mm,width=2.5mm]}] (t13) -- (p8);

\draw[-{Latex[length=2mm,width=2.5mm]}] (t8) -- (p7 |- t8) -- (p7);

\draw[-{Latex[length=2mm,width=2.5mm]}] (p8) -- (t14);
\draw[-{Latex[length=2mm,width=2.5mm]}] (t14) -- (p9);

\draw[-{Latex[length=2mm,width=2.5mm]}] (p8) -- (t15);
\draw[-{Latex[length=2mm,width=2.5mm]}] (t15) -- (p8);
\end{tikzpicture}

%% file: tables/alignments.tex

\begin{table}[ht!]
  \centering
  \setlength{\tabcolsep}{3pt}
  \begin{tabular}{c|c|c|c|c|c|c|c|c|}
  \cline{2-9}
  \multirow{2}{*}{$\sigma_1=$} &
  {$\tau$} & {$e_8$} & {$\gg$} & {$e_6$} & {$e_2$} & {$e_1$} & {$e_1$} & {$e_9$} \\ 
  \cmidrule[1pt]{2-9}
  &
  {$M_1$} & {$e_8$} & {$e_8$} & {$\gg$} & {$\gg$} & {$\gg$} & {$\gg$} & {$\gg$} \\ 
  \cline{2-9}
  \end{tabular}
\vspace{8pt}

  \begin{tabular}{c|c|c|c|c|c|c|c|c|c|c|}
  \cline{2-11}
  \multirow{2}{*}{$\sigma_2=$} &
  {$\tau$} & {$e_8$} & {$e_6$} & {$e_2$} & {$e_1$} & {$e_1$} & {$e_9$} & {$\gg$} & {$\gg$} & {$\gg$}\\ 
  \cmidrule[1pt]{2-11}
  &
  {$M_2$} & {$e_8$} & {$e_6$} & {$\gg$} & {$e_1$} & {$e_1$} & {$e_9$} & {$e_4$} & {$e_3$} & {$e_3$} \\ 
  \cline{2-11}
  \end{tabular}
  \label{tb:optalign}
\end{table}

%% file: tex/evaluation.tex
\section{Evaluation}
\label{sec:evaluation}
\subsection{Experiment}
\label{subsec:experiment}

We collected data on ten subjects, each executing 30 traces to each of the three target poses, a relatively small dataset.
Thus, we use cross-validation to appropriately split the available experimental data into training and testing sets.
In the cross-validation, we conduct experiments for 30 iterations.
In each iteration, three traces, one trace per target pose, are selected as testing traces, while the remaining 87 traces, 3 goals $\times$ (30-1) traces, are used for training.
Hence, we test each trace once, resulting in 90 training sets for each subject.

The recognition performance is measured by precision and recall on a micro level.
Since the PM-based GR system may infer multiple possible poses, we follow the method from our previous work~\cite{AIJ_Su} to calculate the precision ($p$) and recall ($r$) for each problem instance and then compute the averages of these individual precision and recall measurements over all the instances.
In each GR instance, where the number of true target poses is one, precision ($p$) is defined as the fraction of correctly inferred target poses (either zero or one) among all the inferred poses (T1, T2, or T3).
Recall ($r$) is defined as the fraction of the correctly inferred true poses. Thus, recall is one if the unique true pose is inferred, regardless of the number of inferred false target poses, and zero otherwise.

The configurable parameters in the PM-based target pose recognition system, mentioned in \cref{sec:approach}, are optimized through a combination of brute force search method and the PRIM algorithm~\cite{friedman1999bump} with Latin hypercube sampling~\cite{helton2003latin}.
The number of selected features $N_f$ is chosen from a set $N_f \in \{1, 2, \ldots, 47\}$, and the number of clusters $N_c$ for event discretization is selected from a set $N_c \in \{10, 20, ..., 200\}$.
We conduct experiments on every combination of $N_f$ and $N_c$, and then select the $N_f$ and $N_c$ values that result in the highest recognition performance (F1-score defined by $\nicefrac{\left(2 p \,\times\, r\right)}{\left(p+r\right)}$).
For parameters such as $\phi$, $\delta$, and $\lambda$ in \cref{eq:omega}, we followed the method described in \cite{AIJ_Su} to utilize the PRIM algorithm and explored 100 different parameter configurations, and selected the configuration that yielded the highest F1-score.

\subsection{Baselines}
\label{subsec:baselines}
We compare the performance of the PM-based GR system with two state-of-the-art target pose recognition techniques based on LSTM and LDA (cf. \cref{sec:related:work} for details).
The LSTM neural network is specifically designed to capture dependencies and patterns in sequential data, making it suitable for classifying multi-dimensional, real-valued, continuous measurements (sEMG and kinematic signals) to identify the target pose of the prosthesis.
We implement the LSTM network using the same settings and hyperparameters outlined in~
\cite{huang2021development} as a benchmark for comparison.
LDA is a trainable classification function that linearly separates multi-dimensional, real-valued, continuous measurements (data points) into a specified number of clusters.
However, it is typically applied to classify single data points, such as the sEMG and kinematic signals at a specific timestamp, rather than a sequence of signals.
In this transhumeral prosthesis scenario, where the instantaneous sEMG and kinematic signals at different target poses are distinct, the LDA classifier is trained with sets of signals collected during the period when the arm is held at three different target poses.
When testing the recognition performance of LDA, as the prosthesis moves closer and closer to a specific target pose, the last observed instantaneous signals become increasingly recognizable to the trained LDA classifier.
Therefore, we implement an existing LDA-based target pose recognition approach~\cite{DBLP:conf/smc/YuG00CO21} based on the last instantaneous signal of an observed trace, rather than a sequence of signals like the PM or LSTM approaches.

All recognition approaches (PM, LSTM, LDA) were executed on a single core of an Intel Xeon Processor@2.0GHz, using the same selected features and employing the same cross-validation strategy to split the experimental dataset into training and testing sets.
The PM and LSTM approaches were trained using the sequences of signals collected during the arm movement towards target poses, while the LDA approach was trained using 10 data points collected after the arm reaches the target pose and holds at the final position. These signals used to train the LDA approach are not used to train the PM and LSTM approaches.
As GR techniques aim to identify goals before the full sequences of signals are observed, we evaluate the approaches using prefixes of the full traces that are cut off at the first 10\%, 30\%, 50\%, and 70\% of the full sequences.

\subsection{Results}
\label{subsec:results}

The average precisions ($p$) and recalls ($r$), obtained by experimenting on each individual subject to recognize the target poses using different lengths of prefixes (10\%, 30\%, 50\%, and 70\% of the full trace), are presented in \cref{tab:detailed_results}.
The two columns ``Features'' and ``Clusters'' show the number of selected features, $N_f$, and the number of discrete event clusters, $N_c$, for each subject.
Since the LSTM- and LDA-based approaches only infer one possible pose for each problem instance, the precision and recall values for these approaches remain the same in all cases.
The last row shows the average across all subjects and all levels of observation.

\input{tables/detailed_results.tex}

We performed t-tests to examine the statistical significance of the differences in average precisions and recalls between the PM-based approach and the other two benchmarks.
The null hypothesis of the t-tests is that there is no significant difference between the average precisions and recalls.
The t-tests comparing the average precision and recall between the PM- and LSTM-based approaches yield p-values of \mbox{1.185e-03} and \mbox{4.173e-34}, respectively. Similarly, the p-values between the PM- and LDA-based approaches are \mbox{5.153e-10} and \mbox{1.188e-34}.
The results from the t-tests indicate that the average precisions and recalls for different approaches are significantly different from each other at a 95\% confidence level (\v{S}id\'ak correction considered).
Therefore, on average, the PM-based GR approach significantly outperforms the LSTM and LDA-based approaches in terms of precision and recall.
Furthermore, in situations where the system makes mistakes, we analyze the probability gaps, which are defined as the difference between the maximum probability associated with the inferred goals and the probability associated with the true goal.
We observe the following average probability gaps: 0.064 for the PM-based system, 0.277 for the LSTM-based approach, and 0.623 for the LDA-based approach. These results indicate that the PM-based GR system exhibits less confidence when making mistakes than the other two benchmarks.

\Cref{tab:each_subject} shows the average precisions and recalls, along with the 95\% confidence intervals, across all levels of observation for each subject.
Differences in performance are subject-dependent.
Subjects 5 and 10 lead to a broader performance gap between all approaches. 
Still, the PM-based approach outperforms other methods in eight out of ten subjects. 

\input{tables/each_subject.tex}

\Cref{tab:each_obs} shows the average precisions and recalls across all ten subjects for different levels of observation (10\%, 30\%, 50\%, and 70\%), along with the corresponding 95\% confidence intervals.
The precision of all approaches shows a gradual increase as more signals are observed. However, for the PM-based approach, the recall initially starts high, then slightly drops, and finally increases slightly again.
The PM-based approach exhibits significantly higher recall than the other approaches at all levels of observation. For precision, the PM-based approach is comparable to the other approaches at 10\% and 30\%, and significantly higher at 50\% and 70\% levels of observation.
Note that LDA performs well when the full trace is observed, achieving a precision ($p$) and recall ($r$) of $p=r=0.709\pm0.030$. In comparison, the PM approach achieves a precision of $p=0.645\pm0.026$ and a recall of $r=0.793\pm0.027$. This is due to the high similarity of LDA's training data with the observed signal at the end of the trace, when the subject already achieved the target pose.

\input{tables/each_obs.tex}

%% file: tables/detailed_results.tex
\begin{table}[t]
    \resizebox{0.48\textwidth}{!}{%
		\huge
    \begin{tabular}{|cccc|cc|c|c|}
      \hline
      \multicolumn{1}{|c|}{\multirow{2}{*}{Subject}} &
        \multicolumn{1}{c|}{\multirow{2}{*}{Features}} &
        \multicolumn{1}{c|}{\multirow{2}{*}{Clusters}} &
        \multirow{2}{*}{Obs\%} &
        \multicolumn{2}{c|}{PM} &
        LSTM &
        LDA \\ \cline{5-8} 
      \multicolumn{1}{|c|}{} &
        \multicolumn{1}{c|}{} &
        \multicolumn{1}{c|}{} &
         &
        \multicolumn{1}{c|}{$p$} &
        $r$ &
        $p = r$ &
        $p = r$ \\ \hline
      \multicolumn{1}{|c|}{\multirow{4}{*}{1}} &
        \multicolumn{1}{c|}{\multirow{4}{*}{29}} &
        \multicolumn{1}{c|}{\multirow{4}{*}{70}} &
        10\% &
        \multicolumn{1}{c|}{0.365} &
        \textbf{0.967} &
        0.389 &
        \textbf{0.411} \\ \cline{4-8} 
      \multicolumn{1}{|c|}{} &
        \multicolumn{1}{c|}{} &
        \multicolumn{1}{c|}{} &
        30\% &
        \multicolumn{1}{c|}{\textbf{0.431}} &
        \textbf{0.722} &
        0.422 &
        0.378 \\ \cline{4-8} 
      \multicolumn{1}{|c|}{} &
        \multicolumn{1}{c|}{} &
        \multicolumn{1}{c|}{} &
        50\% &
        \multicolumn{1}{c|}{0.491} &
        \textbf{0.744} &
        \textbf{0.500} &
        0.456 \\ \cline{4-8} 
      \multicolumn{1}{|c|}{} &
        \multicolumn{1}{c|}{} &
        \multicolumn{1}{c|}{} &
        70\% &
        \multicolumn{1}{c|}{\textbf{0.628}} &
        \textbf{0.822} &
        0.589 &
        0.611 \\ \hline
      \multicolumn{1}{|c|}{\multirow{4}{*}{2}} &
        \multicolumn{1}{c|}{\multirow{4}{*}{1}} &
        \multicolumn{1}{c|}{\multirow{4}{*}{10}} &
        10\% &
        \multicolumn{1}{c|}{\textbf{0.328}} &
        \textbf{0.967} &
        0.300 &
        0.289 \\ \cline{4-8} 
      \multicolumn{1}{|c|}{} &
        \multicolumn{1}{c|}{} &
        \multicolumn{1}{c|}{} &
        30\% &
        \multicolumn{1}{c|}{\textbf{0.361}} &
        \textbf{0.911} &
        0.333 &
        0.333 \\ \cline{4-8} 
      \multicolumn{1}{|c|}{} &
        \multicolumn{1}{c|}{} &
        \multicolumn{1}{c|}{} &
        50\% &
        \multicolumn{1}{c|}{\textbf{0.396}} &
        \textbf{0.967} &
        0.333 &
        0.244 \\ \cline{4-8} 
      \multicolumn{1}{|c|}{} &
        \multicolumn{1}{c|}{} &
        \multicolumn{1}{c|}{} &
        70\% &
        \multicolumn{1}{c|}{\textbf{0.435}} &
        \textbf{0.956} &
        0.333 &
        0.311 \\ \hline
      \multicolumn{1}{|c|}{\multirow{4}{*}{3}} &
        \multicolumn{1}{c|}{\multirow{4}{*}{2}} &
        \multicolumn{1}{c|}{\multirow{4}{*}{150}} &
        10\% &
        \multicolumn{1}{c|}{0.328} &
        \textbf{0.978} &
        \textbf{0.333} &
        0.322 \\ \cline{4-8} 
      \multicolumn{1}{|c|}{} &
        \multicolumn{1}{c|}{} &
        \multicolumn{1}{c|}{} &
        30\% &
        \multicolumn{1}{c|}{\textbf{0.446}} &
        \textbf{0.878} &
        0.322 &
        0.344 \\ \cline{4-8} 
      \multicolumn{1}{|c|}{} &
        \multicolumn{1}{c|}{} &
        \multicolumn{1}{c|}{} &
        50\% &
        \multicolumn{1}{c|}{\textbf{0.524}} &
        \textbf{0.778} &
        0.311 &
        0.333 \\ \cline{4-8} 
      \multicolumn{1}{|c|}{} &
        \multicolumn{1}{c|}{} &
        \multicolumn{1}{c|}{} &
        70\% &
        \multicolumn{1}{c|}{\textbf{0.606}} &
        \textbf{0.744} &
        0.333 &
        0.333 \\ \hline
      \multicolumn{1}{|c|}{\multirow{4}{*}{4}} &
        \multicolumn{1}{c|}{\multirow{4}{*}{34}} &
        \multicolumn{1}{c|}{\multirow{4}{*}{50}} &
        10\% &
        \multicolumn{1}{c|}{0.344} &
        \textbf{0.956} &
        0.322 &
        \textbf{0.356} \\ \cline{4-8} 
      \multicolumn{1}{|c|}{} &
        \multicolumn{1}{c|}{} &
        \multicolumn{1}{c|}{} &
        30\% &
        \multicolumn{1}{c|}{0.352} &
        \textbf{0.911} &
        0.422 &
        \textbf{0.433} \\ \cline{4-8} 
      \multicolumn{1}{|c|}{} &
        \multicolumn{1}{c|}{} &
        \multicolumn{1}{c|}{} &
        50\% &
        \multicolumn{1}{c|}{0.387} &
        \textbf{0.856} &
        0.433 &
        \textbf{0.444} \\ \cline{4-8} 
      \multicolumn{1}{|c|}{} &
        \multicolumn{1}{c|}{} &
        \multicolumn{1}{c|}{} &
        70\% &
        \multicolumn{1}{c|}{0.480} &
        \textbf{0.833} &
        0.533 &
        \textbf{0.556} \\ \hline
      \multicolumn{1}{|c|}{\multirow{4}{*}{5}} &
        \multicolumn{1}{c|}{\multirow{4}{*}{32}} &
        \multicolumn{1}{c|}{\multirow{4}{*}{90}} &
        10\% &
        \multicolumn{1}{c|}{\textbf{0.356}} &
        \textbf{0.956} &
        \textbf{0.356} &
        0.244 \\ \cline{4-8} 
      \multicolumn{1}{|c|}{} &
        \multicolumn{1}{c|}{} &
        \multicolumn{1}{c|}{} &
        30\% &
        \multicolumn{1}{c|}{\textbf{0.509}} &
        \textbf{0.856} &
        0.478 &
        0.278 \\ \cline{4-8} 
      \multicolumn{1}{|c|}{} &
        \multicolumn{1}{c|}{} &
        \multicolumn{1}{c|}{} &
        50\% &
        \multicolumn{1}{c|}{\textbf{0.572}} &
        \textbf{0.756} &
        0.456 &
        0.211 \\ \cline{4-8} 
      \multicolumn{1}{|c|}{} &
        \multicolumn{1}{c|}{} &
        \multicolumn{1}{c|}{} &
        70\% &
        \multicolumn{1}{c|}{\textbf{0.789}} &
        \textbf{0.856} &
        0.489 &
        0.422 \\ \hline
      \multicolumn{1}{|c|}{\multirow{4}{*}{6}} &
        \multicolumn{1}{c|}{\multirow{4}{*}{28}} &
        \multicolumn{1}{c|}{\multirow{4}{*}{160}} &
        10\% &
        \multicolumn{1}{c|}{0.339} &
        \textbf{0.911} &
        \textbf{0.367} &
        \textbf{0.367} \\ \cline{4-8} 
      \multicolumn{1}{|c|}{} &
        \multicolumn{1}{c|}{} &
        \multicolumn{1}{c|}{} &
        30\% &
        \multicolumn{1}{c|}{\textbf{0.406}} &
        \textbf{0.722} &
        0.289 &
        0.389 \\ \cline{4-8} 
      \multicolumn{1}{|c|}{} &
        \multicolumn{1}{c|}{} &
        \multicolumn{1}{c|}{} &
        50\% &
        \multicolumn{1}{c|}{\textbf{0.454}} &
        \textbf{0.600} &
        0.333 &
        0.344 \\ \cline{4-8} 
      \multicolumn{1}{|c|}{} &
        \multicolumn{1}{c|}{} &
        \multicolumn{1}{c|}{} &
        70\% &
        \multicolumn{1}{c|}{0.569} &
        \textbf{0.667} &
        0.467 &
        \textbf{0.611} \\ \hline
      \multicolumn{1}{|c|}{\multirow{4}{*}{7}} &
        \multicolumn{1}{c|}{\multirow{4}{*}{22}} &
        \multicolumn{1}{c|}{\multirow{4}{*}{80}} &
        10\% &
        \multicolumn{1}{c|}{\textbf{0.387}} &
        \textbf{0.911} &
        0.356 &
        0.356 \\ \cline{4-8} 
      \multicolumn{1}{|c|}{} &
        \multicolumn{1}{c|}{} &
        \multicolumn{1}{c|}{} &
        30\% &
        \multicolumn{1}{c|}{\textbf{0.391}} &
        \textbf{0.744} &
        0.344 &
        0.322 \\ \cline{4-8} 
      \multicolumn{1}{|c|}{} &
        \multicolumn{1}{c|}{} &
        \multicolumn{1}{c|}{} &
        50\% &
        \multicolumn{1}{c|}{0.443} &
        \textbf{0.700} &
        0.444 &
        \textbf{0.467} \\ \cline{4-8} 
      \multicolumn{1}{|c|}{} &
        \multicolumn{1}{c|}{} &
        \multicolumn{1}{c|}{} &
        70\% &
        \multicolumn{1}{c|}{0.480} &
        \textbf{0.733} &
        0.489 &
        \textbf{0.544} \\ \hline
      \multicolumn{1}{|c|}{\multirow{4}{*}{8}} &
        \multicolumn{1}{c|}{\multirow{4}{*}{34}} &
        \multicolumn{1}{c|}{\multirow{4}{*}{100}} &
        10\% &
        \multicolumn{1}{c|}{0.322} &
        \textbf{0.822} &
        \textbf{0.378} &
        \textbf{0.378} \\ \cline{4-8} 
      \multicolumn{1}{|c|}{} &
        \multicolumn{1}{c|}{} &
        \multicolumn{1}{c|}{} &
        30\% &
        \multicolumn{1}{c|}{\textbf{0.409}} &
        \textbf{0.678} &
        0.389 &
        0.400 \\ \cline{4-8} 
      \multicolumn{1}{|c|}{} &
        \multicolumn{1}{c|}{} &
        \multicolumn{1}{c|}{} &
        50\% &
        \multicolumn{1}{c|}{0.522} &
        \textbf{0.678} &
        \textbf{0.533} &
        0.411 \\ \cline{4-8} 
      \multicolumn{1}{|c|}{} &
        \multicolumn{1}{c|}{} &
        \multicolumn{1}{c|}{} &
        70\% &
        \multicolumn{1}{c|}{0.689} &
        \textbf{0.744} &
        0.522 &
        \textbf{0.700} \\ \hline
      \multicolumn{1}{|c|}{\multirow{4}{*}{9}} &
        \multicolumn{1}{c|}{\multirow{4}{*}{23}} &
        \multicolumn{1}{c|}{\multirow{4}{*}{100}} &
        10\% &
        \multicolumn{1}{c|}{0.333} &
        \textbf{0.989} &
        0.333 &
        \textbf{0.378} \\ \cline{4-8} 
      \multicolumn{1}{|c|}{} &
        \multicolumn{1}{c|}{} &
        \multicolumn{1}{c|}{} &
        30\% &
        \multicolumn{1}{c|}{\textbf{0.393}} &
        \textbf{0.878} &
        0.389 &
        0.344 \\ \cline{4-8} 
      \multicolumn{1}{|c|}{} &
        \multicolumn{1}{c|}{} &
        \multicolumn{1}{c|}{} &
        50\% &
        \multicolumn{1}{c|}{0.383} &
        \textbf{0.633} &
        \textbf{0.411} &
        0.344 \\ \cline{4-8} 
      \multicolumn{1}{|c|}{} &
        \multicolumn{1}{c|}{} &
        \multicolumn{1}{c|}{} &
        70\% &
        \multicolumn{1}{c|}{0.398} &
        \textbf{0.622} &
        \textbf{0.478} &
        0.556 \\ \hline
      \multicolumn{1}{|c|}{\multirow{4}{*}{10}} &
        \multicolumn{1}{c|}{\multirow{4}{*}{28}} &
        \multicolumn{1}{c|}{\multirow{4}{*}{170}} &
        10\% &
        \multicolumn{1}{c|}{0.333} &
        \textbf{1.000} &
        \textbf{0.344} &
        0.256 \\ \cline{4-8} 
      \multicolumn{1}{|c|}{} &
        \multicolumn{1}{c|}{} &
        \multicolumn{1}{c|}{} &
        30\% &
        \multicolumn{1}{c|}{\textbf{0.567}} &
        \textbf{0.833} &
        0.511 &
        0.233 \\ \cline{4-8} 
      \multicolumn{1}{|c|}{} &
        \multicolumn{1}{c|}{} &
        \multicolumn{1}{c|}{} &
        50\% &
        \multicolumn{1}{c|}{\textbf{0.685}} &
        \textbf{0.833} &
        0.511 &
        0.278 \\ \cline{4-8} 
      \multicolumn{1}{|c|}{} &
        \multicolumn{1}{c|}{} &
        \multicolumn{1}{c|}{} &
        70\% &
        \multicolumn{1}{c|}{\textbf{0.844}} &
        \textbf{0.911} &
        0.611 &
        0.589 \\ \hline
      \multicolumn{4}{|c|}{Average} &
        \multicolumn{1}{c|}{$\textbf{0.462}\pm0.041$} &
        $\textbf{0.826}\pm0.037$ &
        $0.412\pm0.027$ &
        $0.389\pm0.037$ \\ \hline
    \end{tabular}%
    }
    \caption{Average precision ($p$) and recall ($r$) for individual subjects at different levels of observation (highest in bold).} 
    \label{tab:detailed_results}
		\vspace{-5mm}
\end{table}

%% file: tables/each_subject.tex
\begin{table}[ht!]
    \centering
    \resizebox{0.48\textwidth}{!}{%
    \begin{tabular}{|c|cc|c|c|}
        \hline
        \multirow{2}{*}{Subject} & \multicolumn{2}{c|}{PM}                                & LSTM            & LDA             \\ \cline{2-5} 
                                 & \multicolumn{1}{c|}{$p$}               & $r$               & $p = r$             & $p = r$             \\ \hline
        1                        & \multicolumn{1}{c|}{\textbf{0.479}$~\pm~$0.035} & \textbf{0.814}$~\pm~$0.040 & 0.475$~\pm~$0.052 & 0.464$~\pm~$0.052 \\ \hline
        2                        & \multicolumn{1}{c|}{\textbf{0.380}$~\pm~$0.020} & \textbf{0.950}$~\pm~$0.023 & 0.325$~\pm~$0.049 & 0.294$~\pm~$0.047 \\ \hline
        3                        & \multicolumn{1}{c|}{\textbf{0.476}$~\pm~$0.034} & \textbf{0.844}$~\pm~$0.038 & 0.325$~\pm~$0.049 & 0.333$~\pm~$0.049 \\ \hline
        4                        & \multicolumn{1}{c|}{0.391$~\pm~$0.023} & \textbf{0.889}$~\pm~$0.033 & 0.428$~\pm~$0.051 & \textbf{0.447}$~\pm~$0.052 \\ \hline
        5                        & \multicolumn{1}{c|}{\textbf{0.556}$~\pm~$0.037} & \textbf{0.856}$~\pm~$0.036 & 0.444$~\pm~$0.052 & 0.289$~\pm~$0.047 \\ \hline
        6                        & \multicolumn{1}{c|}{\textbf{0.442}$~\pm~$0.039} & \textbf{0.725}$~\pm~$0.046 & 0.364$~\pm~$0.050 & 0.428$~\pm~$0.051 \\ \hline
        7                        & \multicolumn{1}{c|}{\textbf{0.425}$~\pm~$0.034} & \textbf{0.772}$~\pm~$0.044 & 0.408$~\pm~$0.051 & 0.422$~\pm~$0.051 \\ \hline
        8                        & \multicolumn{1}{c|}{\textbf{0.486}$~\pm~$0.040} & \textbf{0.731}$~\pm~$0.046 & 0.456$~\pm~$0.052 & 0.472$~\pm~$0.052 \\ \hline
        9                        & \multicolumn{1}{c|}{0.377$~\pm~$0.030} & \textbf{0.781}$~\pm~$0.043 & 0.403$~\pm~$0.051 & \textbf{0.406}$~\pm~$0.051 \\ \hline
        10 & \multicolumn{1}{c|}{\textbf{0.607}$~\pm~$0.037} & \textbf{0.894}$~\pm~$0.032 & 0.494$~\pm~$0.052 & 0.339$~\pm~$0.049 \\ \hline
    \end{tabular}%
    }
    \caption{Average precision ($p$) and recall ($r$) at all levels of observation for ten individual subjects (highest in bold).} 
    \label{tab:each_subject}
		\vspace{-3mm}
\end{table}

%% file: tables/each_obs.tex
\begin{table}[h!]
\centering
    \resizebox{0.48\textwidth}{!}{%
    \begin{tabular}{|c|cc|c|c|}
        \hline
        \multirow{2}{*}{Obs\%} & \multicolumn{2}{c|}{PM}                                & LSTM            & LDA             \\ \cline{2-5} 
                               & \multicolumn{1}{c|}{$p$}               & $r$               & $p = r$             & $p = r$             \\ \hline
        10\%                   & \multicolumn{1}{c|}{0.344$~\pm~$0.009} & \textbf{0.946}$~\pm~$0.015 & \textbf{0.348}$~\pm~$0.031 & 0.336$~\pm~$0.031 \\ \hline
        30\%                   & \multicolumn{1}{c|}{\textbf{0.426}$~\pm~$0.020} & \textbf{0.813}$~\pm~$0.026 & 0.390$~\pm~$0.032 & 0.346$~\pm~$0.031 \\ \hline
        50\%                   & \multicolumn{1}{c|}{\textbf{0.486}$~\pm~$0.024} & \textbf{0.754}$~\pm~$0.028 & 0.427$~\pm~$0.032 & 0.353$~\pm~$0.031 \\ \hline
        70\%                   & \multicolumn{1}{c|}{\textbf{0.592}$~\pm~$0.026} & \textbf{0.789}$~\pm~$0.027 & 0.484$~\pm~$0.033 & 0.523$~\pm~$0.033 \\ \hline
    \end{tabular}%
    }
    \caption{Average precision ($p$) and recall ($r$) for all subjects at 10\%, 30\%, 50\% and 70\% observation (highest in bold).} 
    \label{tab:each_obs}
		\vspace{-3mm}
\end{table}

%% file: tex/discussion.tex
\vspace{2mm}
\section{Limitations \& Future Work}\label{sec:discussion}

While our work provides a novel and promising approach to goal recognition for transhumeral prostheses, several aspects can be further elaborated.
Although we used existing datasets for comparison with the state-of-the-art, real-world deployment may require more extensive experiments, including a larger number of subjects (currently 10) and a larger number of (and more complex) target poses and goals (currently three).

Our experiments confirmed that feature selection and discretization have a significant impact on the accuracy of the system (in our case, of the GR outcome). Our current approach is arguably simple and we would like to explore elaborate feature selection and event discretization. For example, it would be interesting to identify feature conditions capturing \emph{meaningful} prosthetic postures/configurations, and test whether sequences of these yield good predictor for the goal being pursued. Doing so will also allow us to take advantage of one of the key features of our PM-based GR approach, namely, the possibility to explain the outcome of the system based on the process model and misalignment of the observed behavior.

Yet another area for further exploration is the testing of alternatives alignment approaches. While process discovery can be done in linear time~\cite{DBLP:conf/icpm/LeemansPW19}, alignment techniques---including the one we used in this work~\cite{DBLP:journals/widm/AalstAD12}---are often exponential in the worst case.
Since our approach does require alignments to be computed at recognition time, extracting them as fast as possible is important for real-time applicability. However, the worst cases often do not come up in practice, and our alignment system took around $0.04$ seconds per goal (note alignments across goals can be parallelized). Nonetheless, we would like to experiment with techniques that are specifically designed for \emph{online} conformance checking~\cite{Zelst.etal:JDSA19} as well as those that seek \emph{approximate} alignments, but \emph{fast}~\cite{vanDongen:AISE17}. Indeed, we conjecture online conformance checking approaches may be a good fit for our GR setting, since observations are incrementally extended and such approaches extract alignments incrementally by re-using previously computed ones.

The used dataset captured the features from forward-reaching tasks using sound limbs rather than real-time control of a prosthesis.
The difference between real-time prosthetic states and sound limbs introduces variations in visual feedback, which could affect the feature patterns collected during movement.
The feature patterns can have an impact on the real-time recognition accuracy of machine-learning-based techniques \cite{Prost:Woodward2019,Prost:Ortiz-Catalan2015}.
Therefore, collecting data from real-time control experiments and pre-defining standard traces toward the goal with distinguishable feature patterns for training the PM-based GR system has the potential to improve accuracy.


%% file: tex/conclusion.tex
\vspace{1.5mm}
\section{Conclusion}
\label{sec:conclusion}

In this paper, we proposed a novel PM-based GR method to recognize a patient's intended target pose when moving an artificial limb, utilizing continuous sensor data.
Our approach transforms the data into discrete events to discover a behavior model and subsequently predict the goal by aligning the observed behavior with the model. 
Experiments on a real-life dataset demonstrate the effectiveness of our proposed GR approach, which significantly outperforms state-of-the-art LSTM-based and LDA-based machine learning methods on both precision and recall. 
In addition, our GR approach is less confident when wrong, which results in a much smoother movement of the prostheses in less certain scenarios.